\documentclass[letterpaper]{article} 
\usepackage{aaai2027}
\usepackage[hyphens]{url}  
\usepackage{graphicx} 
\usepackage{natbib}  
\usepackage{caption} 
\usepackage{algorithm}
\usepackage{algorithmic}

\usepackage{amsmath}
\usepackage{amssymb}
\usepackage{booktabs}
\usepackage{multirow}
\usepackage{makecell}

\usepackage{newfloat}
\usepackage{listings}
\DeclareCaptionStyle{ruled}{labelfont=normalfont,labelsep=colon,strut=off} 
\floatstyle{ruled}
\newfloat{listing}{tb}{lst}{}
\floatname{listing}{Listing}

\usepackage{booktabs}

\title{Learning Holistic Scene Layout against View Shifts in Cross-View Geo-Localization}
\author{
    Chao Yang\textsuperscript{\rm 1},
    Songlian Li\textsuperscript{\rm 1},
    Zhigang Tu\textsuperscript{\rm 1}\corresponding
}
\affiliations{
    \textsuperscript{\rm 1}the State Key Laboratory of Information Engineering in Surveying, Mapping and Remote Sensing, Wuhan University, China\\
}

\begin{document}

\maketitle

\begin{abstract}
Cross-view geo-localization remains challenging due to drastic appearance variations and geometric distortions caused by viewpoint changes. Existing methods primarily rely on discriminative feature learning or explicit geometric alignment, yet often overlook an important observation: humans tend to match scenes by integrating the spatial context of multiple landmarks rather than relying solely on isolated local visual cues. 
Inspired by this, we propose a Spatial Context Modeling (SCM) module, which traverses feature maps along four orthogonal directions and models interactions among the resulting directional representations at each spatial position. By aggregating complementary directional observations, SCM constructs richer scene layout representations.
Since reliable spatial context depends on robust visual features, we introduce a High-Frequency Information Guidance (HFIG) module that extracts stable structural cues from the frequency domain. By emphasizing edges and contours that are less sensitive to appearance variations, HFIG encourages the model to learn more robust representations.
Extensive experiments demonstrate that our method achieves competitive performance on standard cross-view geo-localization benchmarks while exhibiting strong robustness under more challenging scenarios with unknown orientations and limited fields of view (FoVs).
\end{abstract}


\section{Introduction}
Cross-view geo-localization aims to determine the geographic location of a query image by using a geotagged reference image \cite{bansal2011geo, hu2022beyond, mrgeo}. It provides a vision-based alternative for geo-localization in real scenarios and serves as an important complement when GPS signals are unreliable or unavailable. This task holds enticing application potential in areas such as autonomous driving \cite{kim2017satellite}, drone navigation \cite{zheng2020university}, and accurate delivery \cite{li2019cross}. The task typically locates a query image by retrieving the most similar image in a geotagged reference database \cite{toker2021coming, wang2023fine}. However, due to the substantial differences in appearance and geometry shape between images from disparate views, cross-view geo-localization remains a challenging task.

Early efforts \cite{cvm, cai2019ground, sun2019geocapsnet} trained deep networks to pull matched cross-view pairs together in an embedding space while pushing dissimilar ones apart. To bridge the cross-view domain gap, preprocessing strategies such as polar transformation have been introduced to geometrically align aerial and ground views \cite{SAFA, dsm, seh, lisonglian}. While effective in reducing appearance discrepancies, it introduces visual distortions and critically assumes the ground camera sits at the aerial image center, which rarely holds in practice \cite{vigor, transgeo}. Other methods either impose priors that necessitate complex preprocessing pipelines and restrictive assumptions or resort to intricate designs that incur substantial computational overhead. More fundamentally, these methods often focus on discriminative features or explicit geometric alignment, while omitting the complementary spatial context formed by multiple scene elements, which provides more stable cues for cross-view matching.

\begin{figure*}[!t]
    \centering
    \includegraphics[width=\linewidth]{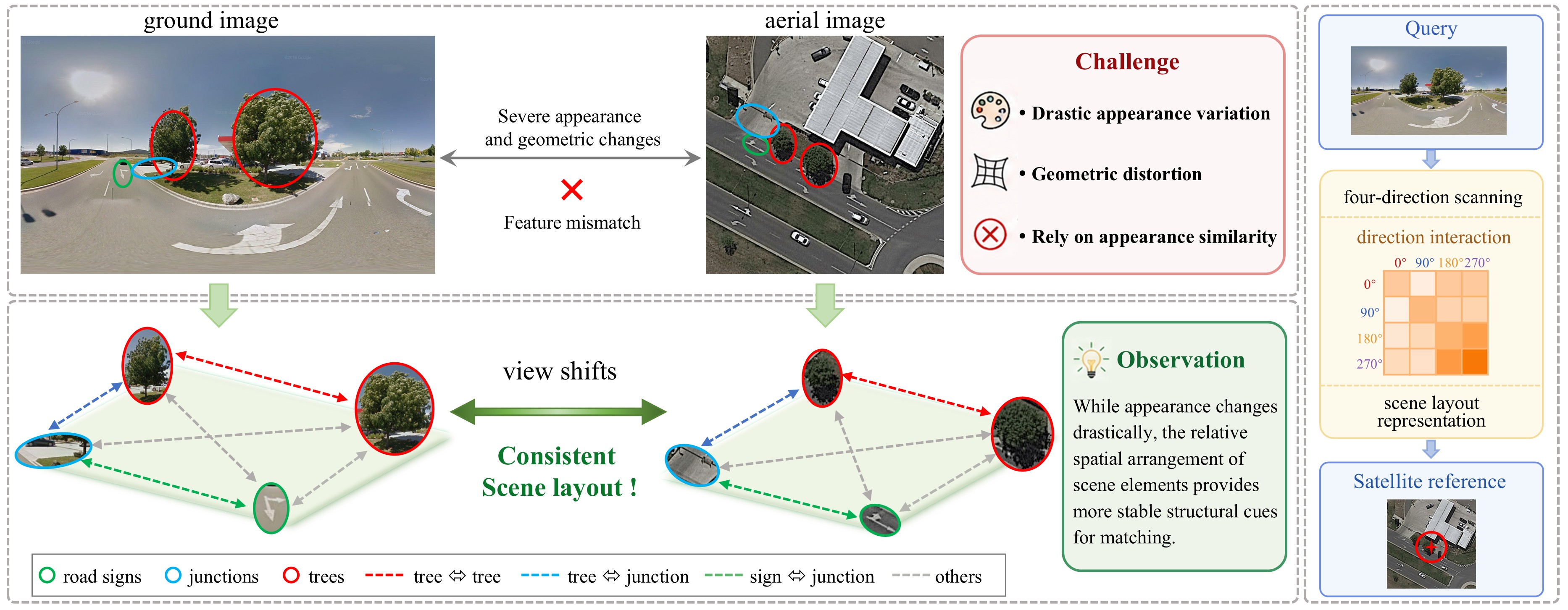}
    \caption{Illustration of the motivation behind our method. Existing methods mainly rely on discriminative appearance features, which will lead to misalignment due to view shifts. When viewpoints change, the relative spatial arrangement of scene elements provides more stable cues than isolated appearance features. Our method aggregates complementary observations from four scanning directions to build richer scene layout representations for robust cross-view alignment.}
    \label{fig1}
\end{figure*}

People do not rely solely on isolated ground objects during visual matching; instead, humans naturally integrate the spatial arrangement of multiple scene elements to infer the overall scene layout, enabling robust matching despite viewpoint changes.
For instance, as shown in Figure~\ref{fig1}, matching based solely on individual objects (e.g., a road sign or a tree) is often ambiguous or insufficient. Instead, incorporating their spatial structure (e.g., ``the sign is adjacent to a junction, which is next to a tree''), that is, the holistic spatial layout of the scene, enables robust and unambiguous cross-view alignment.
However, the effectiveness of spatial layout modeling depends on reliable feature representations. The underlying features extracted in the spatial domain, where features are easily corrupted by illumination shifts and seasonal changes, are unstable. High-frequency components mainly capture structural information such as edges and contours, which are generally less sensitive to appearance variations. By emphasizing stable structural cues in the spatial domain rather than factors susceptible to appearance changes, this frequency information can serve as a reliable auxiliary signal to help achieve cross-view feature alignment. Nevertheless, research on effectively leveraging frequency-domain information for cross-view geo-localization remains scarce. 

Inspired by the insight that humans rely on complementary spatial context rather than isolated visual objects, we propose SLGeo, a framework that integrates multi-directional observations to learn holistic scene layout representations for cross-view geo-localization.
Specifically, we propose a Spatial Context Modeling (SCM) module, which scans feature maps along four orthogonal directions to obtain different directional observations. Instead of relying on a single traversal order, SCM models interactions among the resulting directional representations at each spatial position, enabling each location to integrate context from multiple observation directions.
The complementary directional observations provide richer spatial context and encourage the learning of more holistic scene representations under view shifts.
Moreover, a High-Frequency Information Guidance (HFIG) module is designed to employ the Discrete Cosine Transform (DCT) to transform the image from the spatial domain to the frequency domain, mining high-frequency information, such as contours and edges. These stable structural cues anchor the representation against appearance changes to provide robust structural cues for spatial context modeling.
Extensive experiments on the CVUSA and CVACT demonstrate that SLGeo achieves competitive performance. Furthermore, SLGeo generalizes well to more challenging scenarios involving unknown orientations and limited FoVs, demonstrating strong robustness beyond standard benchmark settings. 
The main contributions are summarized as follows:
\begin{itemize}
    \item We propose a Spatial Context Modeling (SCM) module that aggregates complementary observations from four scanning directions to construct holistic scene layout representations for cross-view geo-localization.
	\item We introduce a High Frequency Information Guided (HFIG) module, which extracts stable structural cues from the frequency domain to facilitate the model in building reliable scene layout representations.
	\item We develop a framework that jointly leverages holistic scene layout modeling and frequency-domain structural guidance for robust cross-view feature alignment, while demonstrating strong generalization under unknown orientations and limited FoVs.
\end{itemize}

\section{Related Work}
\label{sec:related}
\subsection{Feature-based Cross-View Geo-localization}
Early cross-view geo-localization methods relied on hand-crafted features \cite{dalal2005histograms, rublee2011orb, lin2013cross}. 
These methods are difficult to match accurately due to the quality of extracted features.
With the development of deep learning, researchers utilize Siamese or triplet networks to extract discriminative image presentations for cross-view matching \cite{lin2015learning, vo2016localizing}. These methods extract features from ground-view and aerial-view images to bridge the drastic gap between them. 
Hu et al. \cite{cvm} proposed CVM-Net, which integrates the NetVLAD layer \cite{netvlad} to encode global feature descriptors.
CVFT \cite{cvft} transported features from one domain to the other, facilitating feature alignment between ground and aerial images. through explicitly establishing cross-view domain transfer.
L2LTR \cite{l2ltr} exploited the Transformer to learn geometric configurations between ground and aerial images and integrated a self-cross attention mechanism to improve the quality of representations.
TransGeo \cite{transgeo} was a pure transformer-based approach, which proposed an attention-guided non-uniform cropping method to remove uninformative image patches without a negligible drop on performance.
Li et al. \cite{lisonglian} combined multi-scale feature fusion with an attention mechanism, effectively alleviating the matching difficulties caused by scale differences in traditional approaches.
SCPNet \cite{scpnet} used a CNN to extract depth features and proposed a semantic concept mining module to extract semantic concept information.
Despite these feature-based methods having made promising progress, they tend to rely on implicitly modeling spatial information, overlooking the importance of geometric spatial layouts alignment.

\subsection{Geometry-based Cross-View Geo-localization}
Geometry-based methods focus on aligning geometric spatial layout between ground-view and aerial-view images, mitigating feature misalignments. Initial methods use polar transform to convert the aerial view to the ground view, or vice versa, to bridge the appearance gap.
SAFA \cite{SAFA} applied a polar transform to make the aerial image closer to ground-view, and then aggregated features by learning multiple spatial embeddings.
Subsequently, DSM \cite{dsm} learned deep features from the ground and polar-transformed aerial images to obtain the orientation by computing the correlation between them.
While these methods utilizing polar transform have improved performance, they lead to appearance distortion and additional preprocessing, and rely on the assumption that the ground image is located at the center of the aerial image. This assumption has proven to lack general applicability \cite{transgeo}.
PaSS-KD \cite{passkd} provided the local and multi-scale knowledge from ground images as fine-grained location-related supervision in a self-enhanced manner to guide cross-view image feature extraction.
GeoDTR+ \cite{geodtr+} enhanced the geometric layout extractor to fully exploit the information of input features and better model the correlations among visual features, and introduced contrastive hard sample generation to facilitate model training. 

The above methods aim to mitigate the matching ambiguities caused by feature misalignments. Although effective, they mainly rely on feature-level alignment or geometric transformations, paying limited attention to exploiting complementary spatial context for holistic scene understanding.
FRGeo \cite{frgeo} further explores geometric consistency by recombining regional features to explicitly align scene layouts. 
However, its alignment strategy largely relies on reliable directional correspondence and sufficient visual coverage, making it less robust in complex settings.
Instead of explicitly aligning regions or relying on geometric transformations, our method constructs richer spatial context representations by aggregating multiple observation directions to enhance performance. 
Furthermore, compared to FRGeo and Sample4Geo, SLGeo exhibits conspicuous advantages in varying orientation and reduced FoV settings.

\begin{figure*}[t]
  \centering
  \includegraphics[width=\textwidth]{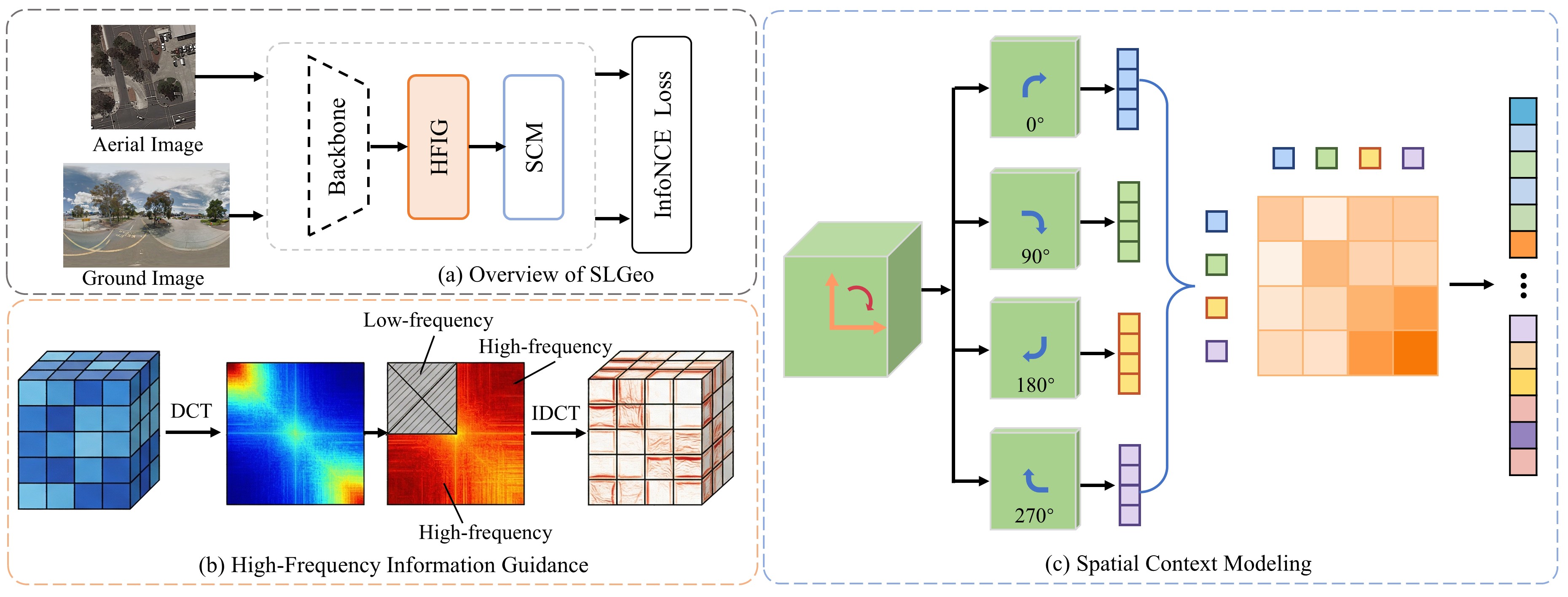}
  \caption{Overview of the proposed method. Illustration of the High-frequency Information Guidance (HFIG) and Spatial Context Modeling (SCM). The HFIG module converts the image into the frequency domain using DCT and then extracts fine-grained information of high-frequency. SCM captures the holistic spatial layout via position-wise multi-head attention over four directional scanning paths.}
  \label{component}
\end{figure*}

\section{Method}
\subsection{Overview}
The proposed model adopts a weight-sharing dual-branch architecture to extract discriminative representations from aerial and ground images. The overall pipeline is illustrated in Figure \ref{component} (a). Given the aerial-view image $I_{a}$ and ground-view image $I_{g}$ pair, the backbone is employed to capture fundamental visual features, obtaining the $F_{a} \in \mathbb{R}^{B \times C \times H_{a} \times W_{a}}$ and $F_{g} \in \mathbb{R}^{B \times C \times H_{g} \times W_{g}}$, respectively. 
On top of the backbone, to improve the ability of the model against viewpoint changes and appearance variations, the High-Frequency Information Guidance (HFIG) module and the Spatial Context Modeling (SCM) module are introduced.
HFIG extracts high-frequency components $\mathbf{F}_h$ from images and injects them back to enhance stable structural cues.
Subsequently, the enhanced features are fed into SCM, which performs four-directional scanning followed by position-wise interaction among directional representations to enhance spatial layout representations.
Finally, the contrastive learning is utilized to optimize the similarity between matched image pairs.

\subsection{High-Frequency Information Guidance}
Existing cross-view geo-localization methods primarily focus on learning discriminative features in the spatial domain. However, these spatial features tend to interfere with factors such as changes in illumination, seasonal color shifts, etc. These inconsistencies in appearance may mislead the model in learning key features, leading to suboptimal matching performance. 
To provide reliable cues for subsequent spatial context aggregation, we propose a High-Frequency Information Guidance (HFIG) module. By transforming features into the frequency domain via the Discrete Cosine Transform (DCT), HFIG emphasizes high-frequency structural information while suppressing low-frequency appearance-sensitive components, thereby providing robust frequency cues for cross-view feature alignment.

Specifically, given the fundamental features $\mathbf{F}$ extracted from the backbone, they are projected from the spatial domain into the frequency domain by utilizing DCT, as shown in Figure \ref{component} (b). Frequency feature $\mathbf{F}_q$ can be represented as:
\begin{equation}
\mathbf{F}_{q}=\mathcal{D}(\mathbf{F}),
\end{equation}
where $\mathcal{D}(\cdot)$ denotes DCT. In this frequency space, the top-left coefficients represent low-frequency components (e.g., global color and luminance), while the remaining regions represent high-frequency details such as edges and contours.

Compared with low-frequency components, high-frequency components preserve structural details and are generally less sensitive to appearance variations. Therefore, a binary mask $\mathbf{M}$ is designed to preserve the high-frequency information while suppressing the low-frequency part:
\begin{equation}
    \mathbf{F}_{H} = \mathbf{F}_q \odot \mathbf{M}, 
    \mathbf{M}(h,w)=
    \begin{cases}
    0,& h<\beta H \ \text{and}\ w<\beta W,\\
    1,& \text{otherwise}.
    \end{cases}
\end{equation}
where $(h,w)$ denote the frequency coordinates, $H, W$ are the size of $\mathbf{F}_{q}$ and $\beta$ is the threshold of mask with a value between $0-1$.
Subsequently, the mined high-frequency features $\mathbf{F}_{H}$ are transformed back to the spatial domain using the inverse DCT. 
Finally, $\mathbf{F}_h$ is adaptively fused with the original spatial feature $\mathbf{F}$. Their formal representations:
\begin{equation}
    \begin{aligned}
            \mathbf{F}_h &= \mathcal{D}^{-1}(\mathbf{F}_{H}),\\
            \mathbf{F}_{HFIG} &= \mathbf{F} + \boldsymbol{\gamma} \cdot \mathbf{F}_h,
    \end{aligned}
\end{equation}
where $\boldsymbol{\gamma}$ is a learnable weight. It allows the network to automatically balance semantic appearance information and frequency-domain structural cues. The enhanced features are subsequently fed into SCM, providing more reliable structural cues for multi-directional spatial context aggregation.

\subsection{Spatial Context Modeling Module}
\label{sec:srm}
The prime challenge of cross-view geo-localization lies in capturing consistent representations across drastic perspective transformations. Although the visual appearance of objects changes drastically across viewpoints, the overall spatial context provides more stable cues for matching than isolated objects.
For example, the spatial relation in scenes, such as ''a junction is next to a sign and a tree is close to another tree," persists even when the viewpoint changes substantially. 
Inspired by the human visual matching process, which relies on a holistic understanding of the spatial layout rather than isolated objects, we propose the Spatial Context Modeling (SCM) module. 
Unlike prior methods that directly model each view into a global descriptor, SCM aggregates complementary directional observations at each spatial position to construct richer spatial context representations.

\noindent\textbf{Multi-Directional Scanning.}
To model the holistic scene layout from multiple orientations, we first traverse the feature map along four rotation scanning ways. Given the frequency-guided feature map $\mathbf{F}_{HFIG} \in \mathbb{R}^{B \times C \times H \times W}$, the RotationScan module generates four complementary sequences by rotating the feature map by $0^\circ$, $90^\circ$, $180^\circ$ and $270^\circ$:
\begin{equation}
    \mathbf{D} = \text{RotationScan}(\mathbf{F}_{HFIG}) \in \mathbb{R}^{B\times4\times C\times L},
\end{equation}
where $L=H\times W$. Specifically,
\begin{equation}
\begin{aligned}
    \mathbf{D}_{1} &= \text{Flatten}(\mathbf{F}_{HFIG}),\\
    \mathbf{D}_{2} &= \text{Flatten}\!\left(\text{Rot}_{90}(\mathbf{F}_{HFIG})\right),\\
    \mathbf{D}_{3} &= \text{Flatten}\!\left(\text{Rot}_{180}(\mathbf{F}_{HFIG})\right),\\
    \mathbf{D}_{4} &= \text{Flatten}\!\left(\text{Rot}_{270}(\mathbf{F}_{HFIG})\right),
\end{aligned}
\end{equation}
where $\text{Rot}_{\theta}(\cdot)$ denotes rotating the feature map by $\theta$ degrees in the spatial plane, and $\text{Flatten}(\cdot)$ denotes flattening.

The RotationScan module provides four orientation sequential representations of the same feature map. Each rotated sequence provides a complementary observation of the same spatial position under a different traversal order, enabling SCM to integrate complementary directional context before feature aggregation.

\noindent\textbf{Position-Wise Direction Interaction.} 
For each spatial position $p \in \{1, \dots, L\}$, its deep features under each of the four scanning directions are collected to form a set of four directional tokens $\{\mathbf{d}_p^{(1)}, \mathbf{d}_p^{(2)}, \mathbf{d}_p^{(3)}, \mathbf{d}_p^{(4)}\} \subset \mathbb{R}^{C}$. Subsequently, a multi-head self-attention over these four tokens is utilized to learn a position-specific Spatial Context Modeling module:
\begin{equation}
    \text{Attn}_p^{(h)} = \text{softmax}\left(\frac{\mathbf{Q}_p^{(h)} \mathbf{K}_p^{(h)\top}}{\sqrt{d}}\right) \in \mathbb{R}^{4 \times 4},
\end{equation}
where $\mathbf{Q}_p^{(h)}, \mathbf{K}_p^{(h)} \in \mathbb{R}^{4 \times d}$ are the query and key for head $h$ at position $p$, $d = C / N_{\text{heads}}$ is the per-head dimension. 
The resulting attention matrix $\mathrm{Attn}_p^{(h)}$ encodes the interactions among the four directional sequences, enabling each spatial position in one scanning sequence to perceive complementary information from the remaining directional sequences. Such cross-direction interaction enriches the spatial context representation at each position and improves robustness to viewpoint and orientation changes. Although the interaction is performed over directional representations, the complementary observations collectively provide a more holistic description of the surrounding scene structure.
Subsequently, $\mathrm{Attn}_p^{(h)}$ is projected back to the original feature dimension:
\begin{equation}
    \mathbf{z}_p = \text{Proj}\left(\text{Concat}_{h=1}^{N_{\text{heads}}} \left(\text{Attn}_p^{(h)} \mathbf{V}_p^{(h)}\right)\right),
\end{equation}
where $\mathbf{V}_p^{(h)}\in\mathbb{R}^{4\times d}$ denotes the value and $\operatorname{Proj}(\cdot)$ is a learnable linear projection initialized with zero weights.

The final feature representations are calculated by:
\begin{equation}
    \mathbf{F}_{SCM} =  \text{$P_{avg}$}\left(\mathbf{z}_p + res \right) \in \mathbb{R}^{B \times C},
\end{equation}
where $P_{avg}(\cdot)$ denotes spatial average pooling and $res$ is the input feature connected through a residual connection.

\subsection{Optimization Objective}
The cross-view geo-localization task aims to project the features of input images into an embedding space, pulling the similar pairs closer and pushing the mismatched pairs far apart. 
Following \cite{sample4geo, congeo}, we adopt a symmetric contrastive learning objective based on the InfoNCE loss. The Loss is defined as:
\begin{equation}
    \mathcal{L}(q, R)_{\text{InfoNCE}} = -\log \frac{\exp(q \cdot r_{+} / \tau)}{\sum_{i=0}^{R} \exp(q \cdot r_{i} / \tau)},
\end{equation}

where $q$ denotes the query ground-view image, $R$ is a set of reference aerial-view images, $r_{+}$ denotes the positive one corresponding to $q$, and $\tau$ is a temperature parameter that controls the difficulty of the loss function.

Moreover, a symmetric optimization strategy is adopted to balance the learning process from both aerial-to-ground and ground-to-aerial directions. The total training objective $\mathcal{L}_{total}$ is the average of the bidirectional losses:
\begin{equation}
    \mathcal{L}_{total} = \frac{1}{2} (\mathcal{L}_{aerial \to ground} + \mathcal{L}_{ground \to aerial}).
\end{equation}

\section{Experiments}
\subsection{Datasets and Evaluation Metrics}
\subsubsection{Datasets.} We evaluate our method on two public cross-view geo-localization datasets.
\textbf{CVUSA} \cite{cvusa} contains 35,532 image pairs for training and 8,884 image pairs for testing. The panorama images are warping to ensure that geographic north is located in the upper centre of the satellite image. This task is a one-to-one mapping, meaning that each satellite image corresponds to one street view image. Referring to previous works \cite{cvusa, SAFA}, we used this subset in our experiments, which contains 35,532 image pairs for training and 8,884 image pairs for testing.
\textbf{CVACT} \cite{cvact} provides 35,532 image pairs for training and 8,884 image pairs for validation, as well as CVUSA. Furthermore, it provides 92,802 image pairs to support fine-grained city-scale geo-localization testing. The task is still a 1 to 1 mapping, and the alignment of images is like in the CVUSA.

\subsubsection{Evaluation Metrics.} Following the literature \cite{cvact, transgeo, frgeo}, we employ top-k recall as the evaluation metric. The metric, denoted as R@K, represents the probability of correct matches among the top-K retrieved results. 

\subsection{Implementation Details}
The model employs a ConvNeXt-Base \cite{convnet} as the backbone and is initialized with off-the-shelf pre-trained weights on ImageNet-1K \cite{imagenet}. Our model is implemented in PyTorch, and all experiments are performed on an NVIDIA GeForce RTX 4090 Server. 
We use the AdamW optimizer with an initial learning rate of $1\times 10^{-4}$ and a cosine learning rate scheduler. The model is trained for 60 epochs with a batch size of 32. Experiments comparing different hyperparameters are provided in the supplementary materials. We use the same training configuration across all datasets.
For CVUSA and CVACT datasets, the image sizes of aerial images and ground images are resized to 384 $\times$ 384 and 140 $\times$ 768, respectively.

\subsection{Comparison}
We compare our method with 11 existing methods on the CVUSA and CVACT datasets, including SAFA \cite{SAFA}, TransGeo \cite{transgeo}, GeoDTR \cite{geodtr}, PaSS-KD \cite{passkd}, Sample4Geo \cite{sample4geo}, Wang et al. \cite{wangetal}, Li et al. \cite{lietal}, FRGeo \cite{frgeo}, ConGeo \cite{congeo}, SCPNet \cite{scpnet}, DSTG \cite{dstg} and SinGeo \cite{singeo}. 
Furthermore, we evaluate the performance of different methods under complex settings such as unknown orientation and limited FoVs on CVUSA and CVACT datasets.

\textbf{Results on CVUSA.} As summarized in Table \ref{tab:cvusa}, note that our method does not employ polar transform, and methods with polar transform are marked with “†”. Nevertheless, SLGeo achieves competitive performance compared to previous works, which highlights the effectiveness of our method in learning the holistic spatial layout. 
Specifically, in the more stringent R@1 and R@5 metrics, our approach achieves the optimal result, outperforming other methods. SLGeo also contains optimal results in the R@10, lagging behind by only 0.02\% in the R@1\%.

\begin{table}[t]
\centering
\resizebox{0.8\linewidth}{!}{
\begin{tabular}{lcccc}
\toprule
\multirow{2}{*}{Method} & \multicolumn{4}{c}{CVUSA} \\
\cmidrule(lr){2-5}
 & R@1 & R@5 & R@10 & R@1\% \\
\midrule

SAFA$\dagger$ & 89.84 & 96.93 & 98.14 & 99.64 \\
TransGeo & 94.08 & 98.36 & 99.04 & 99.77 \\
GeoDTR$\dagger$ & 95.43 & 98.86 & 99.34 & 99.86 \\
PaSS-KD$\dagger$ & 94.09 & 98.42 & 99.13 & 99.77 \\
Sample4Geo & \underline{98.68} & \underline{99.68} & \underline{99.78} & 99.87 \\
Li et al. & 94.88 & 98.80 & 99.36 & 99.77 \\
FRGeo & 97.06 & 99.25 & 99.47 & 99.85 \\
ConGeo & 98.27 & 99.53 & 99.67 & 99.84 \\
SCPNet$\dagger$ & 96.88 & 99.31 & 99.62 & \underline{99.89} \\
DSTG & 95.22 & 98.66 & 99.13 & 99.79 \\
SinGeo & 97.30 & - & - & \textbf{99.90} \\
Ours & \textbf{98.73} & \textbf{99.70} & \textbf{99.81} & 99.88 \\
\bottomrule
\end{tabular}
}
\caption{Performance comparison on CVUSA dataset. Best results are in bold, second-best in underlined. “-” means the score is not provided in the original paper.}
\label{tab:cvusa}
\end{table}

\begin{table}[t]
\centering
\resizebox{\linewidth}{!}{
\small
\setlength{\tabcolsep}{3pt} 
\begin{tabular}{lcccccccc}
\toprule
\multirow{2}{*}{Method} & \multicolumn{4}{c}{CVACT\_val} & \multicolumn{4}{c}{CVACT\_test} \\
\cmidrule(lr){2-5} \cmidrule(lr){6-9}
 & R@1 & R@5 & R@10 & R@1\% & R@1 & R@5 & R@10 & R@1\% \\
\midrule
SAFA$\dagger$ & 81.03 & 92.80 & 94.84 & 98.17 & 55.50 & 79.94 & 85.08 & 94.49 \\
TransGeo & 84.95 & 94.14 & 95.78 & 98.37 & - & - & - & - \\
GeoDTR$\dagger$ & 86.21 & 95.44 & 96.72 & 98.77 & 64.52 & 88.59 & 91.96 & \textbf{98.74} \\
PaSS-KD$\dagger$ & 87.20 & 94.30 & 95.67 & 97.85 & 66.81 & 88.03 & 90.87 & 98.02 \\
Sample4Geo & \underline{90.81} & \underline{96.74} & 97.48 & 98.77 & 71.51 & \underline{92.42} & \underline{94.45} & \underline{98.70} \\
Li et al. & 87.89 & 95.27 & 96.40 & 98.37 & 65.30 & 89.47 & 92.15 & 98.27 \\
FRGeo & 90.35 & 96.45 & 97.25 & 98.74 & \underline{72.15} & 91.93 & 94.05 & 98.66 \\
ConGeo & 90.13 & 96.18 & 97.01 & 98.24 & 71.72 & 91.68 & 93.68 & 98.35 \\
SCPNet$\dagger$ & 90.15 & 96.44 & \textbf{97.69} & \textbf{99.18} & 67.98 & 89.63 & 92.08 & 96.83 \\
DSTG & 86.29 & 95.43 & 96.85 & 98.81 & 61.14 & 87.53 & 91.34 & 98.67 \\
SinGeo & 87.10 & - & - & 96.90 & 69.60 & - & - & 97.10 \\
\textbf{Ours} & \textbf{91.02} & \textbf{96.83} & \underline{97.53} & \underline{98.86} & \textbf{72.28} & \textbf{92.69} & \textbf{94.47} & 98.69 \\
\bottomrule
\end{tabular}
}
\caption{Performance comparison on CVACT dataset. $\dagger$ indicates applying polar transform to aerial images. Best results in bold, second-best in underlined. “-” means the score is not provided in the original paper.}
\label{tab:cvact}
\end{table}

\begin{table*}[t]
\centering
\resizebox{\linewidth}{!}{
\small 
\setlength{\tabcolsep}{3pt} 
\begin{tabular}{c|l|cccc|cccc|cccc|cccc}
\hline
\multirow{2}{*}{Set} & \multirow{2}{*}{Methods} & \multicolumn{4}{c|}{FoV=$360^\circ$} & \multicolumn{4}{c|}{FoV=$180^\circ$} & \multicolumn{4}{c|}{FoV=$90^\circ$} & \multicolumn{4}{c}{FoV=$70^\circ$} \\
 &  & R@1 & R@5 & R@10 & R@1\% & R@1 & R@5 & R@10 & R@1\% & R@1 & R@5 & R@10 & R@1\% & R@1 & R@5 & R@10 & R@1\% \\
\hline
 \multirow{10}{*}{\rotatebox{90}{CVUSA}} 
 & CVM & 16.25 & 38.86 & 49.41 & 88.11 & 7.38 & 22.51 & 32.63 & 75.38 & 2.76 & 10.11 & 16.74 & 55.49  & 
 2.62 & 9.30 & 15.06 & 21.77  \\
 & CVFT & 23.38 & 44.42 & 55.20 & 86.64 & 8.10 & 24.25 & 34.47 & 75.15 & 4.80 & 14.84 & 23.18 & 61.23 & 
 3.79 & 12.44 & 19.33 & 55.56 \\
 & DSM & 78.11 & 89.46 & 92.90 & 98.50 & 48.53 & 68.47 & 75.63 & 93.02 & 16.19 & 31.44 & 39.85 & 71.13 & 8.78 & 19.90 & 27.30 & 61.20 \\
 & SEH & 85.37 & 93.51 & 95.81 & - & 53.73 & 72.32 & 78.96 & - & 16.58 & 32.17 & 40.34 & - & 7.82 & 18.84 & 25.59 & -\\
 & L2LTR & - & - & - & - & 56.69 & 80.86 & 87.75 & 98.01 & 26.92 & 50.49 & 60.41 & 86.88 & 13.95 & 33.07 & 43.86 & 77.65  \\
 & TransGeo & - & - & - & - & 58.22 & 81.33 & 87.66 & 98.13 & 30.12 & 54.18 & 63.96 & 89.18 &  - & - & - & - \\
 & SAIG-D & 71.95 & 90.15 & 94.03 & 99.18 & 52.45 & 78.11 & 85.83 & 97.71 & 26.71 & 50.24 & 59.82 & 86.63 &  20.90 & 41.40 & 51.23 & 80.38 \\
 & Sample4Geo & 90.80 & 97.14 & 98.05 & 99.10 & 83.51 & 94.54 & 96.23 & 98.83 & 47.24 & 71.24 & 79.14 & \textbf{93.96} & 36.77 & 62.20 & 71.92 & 92.06 \\
 & FRGeo & 74.55 & 91.70 & 94.91 & 99.32 & 54.65 & 80.28 & 87.55 & 98.15 & 27.75 & 53.11 & 64.28 & 90.61 & - & - & - & - \\

 & ConGeo & \textbf{96.60} & \textbf{98.90} & \textbf{99.20} & \textbf{99.70} & \underline{92.30} & \underline{97.90} & \textbf{98.70} & \textbf{99.70} & \textbf{55.50} & \textbf{75.40} & \textbf{81.50} & \underline{93.90} & \textbf{49.10} & \textbf{70.80} & \textbf{78.00} & \textbf{93.10} \\

  & Ours & \underline{95.61} & \underline{97.76} & \underline{98.38} & \underline{99.35} & \textbf{92.41} & \textbf{98.01} & \underline{98.61} & \underline{99.67} & \underline{53.68} & \underline{73.60} & \underline{79.87} & 93.01 &
  \underline{43.75} & \underline{67.98} & \underline{75.17} & \underline{92.69} \\

\hline
\multirow{7}{*}{\rotatebox{90}{CVACT}} 
 & CVM & 13.09 & 33.85 & 45.69 & 81.80 & 3.94 & 13.69 & 21.23 & 59.22 & 1.47 & 5.70 & 9.64 & 38.05 & 1.24 & 4.98 & 8.42 & 34.74 \\
 & CVFT & 26.79 & 46.89 & 55.09 & 81.03 & 7.13 & 18.47 & 26.83 & 63.87 & 1.85 & 6.28 & 10.54 & 39.25 & 1.49 & 4.13 & 8.19 & 34.59 \\
 & DSM & 72.91 & 85.70 & 88.88 & 95.28 & 49.12 & 67.83 & 74.18 & 89.93 & 18.11 & 33.34 & 40.94 & 68.65 & 8.29 & 20.72 & 27.13 & 57.08 \\
 & SEH & 77.41 & 88.62 & 90.94 & - & 47.69 & 67.86 & 74.28 & - & 13.85 & 28.39 & 36.24 & - & 6.91 & 16.46 & 22.28 & - \\
 & Sample4Geo & 82.40 & \underline{90.60} & 92.30 & 95.40 & 58.36 & 77.25 & 83.01 & 93.73 & 26.70 & 50.69 & 60.96 & 85.05 & 18.37 & 38.88 & 48.98 & 78.55 \\

 & FRGeo & 55.22 & 79.45 & 85.58 & 96.06 & 10.81 & 26.74 & 36.80 & 70.98 & 9.98 & 25.93 & 35.75 & 71.59 & - & - & - & - \\
 & ConGeo & \underline{83.00} & \underline{90.60} & \underline{92.40} & \underline{96.30} & \underline{70.30} & \textbf{85.20} & \underline{88.60} & \underline{95.10} & \textbf{40.60} & \textbf{62.60} & \textbf{69.80} & \underline{86.60} & \underline{24.60} & \underline{45.30} & \underline{54.30} & \underline{80.60} \\
 & Ours & \textbf{83.17} & \textbf{90.68} & \textbf{92.56} & \textbf{96.53} & \textbf{70.62} & \underline{85.16} & \textbf{88.79} & \textbf{95.35} & \underline{40.28} & \underline{61.99} & \underline{69.00} & \textbf{87.72} &
  \textbf{24.79} & \textbf{45.45} & \textbf{54.42} & \textbf{80.92} \\
\hline
\end{tabular}
}
\caption{Comparison of images with unknown orientation and different FoVs on the CVACT and CVUSA datasets. ‘’-'' means the score is not provided in the original paper. The best performance is in bold and the second-best performance is underline. Note that the model is trained and tested on FoV-specific images.}
\label{tab:fov_comparison}
\end{table*}

\textbf{Results on CVACT.} 
To further evaluate the retrieval performance, we conduct experiments on the CVACT dataset. CVACT provides both a validation set and a more challenging large-scale test set. The results are presented in Table \ref{tab:cvact}.
On CVACT\_val, SLGeo achieves the best R@1 of 91.02\% and R@5 of 96.83\%, respectively. For R@10 and R@1\%, SLGeo ranks surpass Sample4Geo, second behind SCPNet. On the more challenging CVACT\_test, SLGeo ranks within the top one for R@1, R@5 and R@10. For R@1\%, SLGeo is slightly inferior to GeoDTR and Sample4Geo.

\textbf{Generalization with Unknown Orientation and Limited FoVs.} We conduct experiments in a more general and realistic scenario on CVUSA and CVACT datasets to evaluate model robustness, where the orientations of both ground-view and aerial-view images are shifted, and the FoVs of ground-view images are smaller than $360^\circ$. Following prior work, we randomly shift and crop ground images along the azimuth direction with three FoV settings ($360^\circ, 180^\circ, 90^\circ$). The results are summarized in Table \ref{tab:fov_comparison}. As expected, all methods degrade as the FoV narrows due to the loss of spatial context. However, SLGeo demonstrates strong robustness across both datasets and all FoV levels.

On CVUSA, SLGeo consistently ranks among the top two methods across all FoV settings. In particular, it achieves the best performance at FoV=$180^\circ$ and remains highly competitive under more challenging FoV=$90^\circ$ and $70^\circ$ settings, where only ConGeo obtains slightly higher recall. A similar trend is observed on CVACT, where SLGeo achieves the best overall performance in most settings while consistently remaining within the top two across all FoVs.

Notably, compared to other standard cross-view localization methods such as FRGeo and Sample4Geo, our method demonstrates significantly superior robustness in complex scenarios involving unknown orientations and limited FoVs. 
These results suggest that aggregating complementary directional observations becomes increasingly beneficial as the available visual context decreases. When the field of view is severely restricted, relying solely on discriminative features often leads to ambiguous matching. 
In contrast, SCM integrates complementary observations from multiple scanning directions at each spatial position, providing additional spatial context that compensates for the missing visual information, leading to more robust feature representations than methods relying primarily on discriminative features.
Although ConGeo achieves slightly higher performance under several challenging FoV settings, it is specifically designed for such complex conditions. By contrast, SLGeo achieves consistently competitive performance across both standard and challenging scenarios, demonstrating strong robustness.

\subsection{Ablation Study}
\label{sec:ablation}
To evaluate the effectiveness of each proposed component, we conduct ablation experiments on the CVACT dataset using ConvNeXt-Base as the backbone of the baseline model. We incrementally integrate the Spatial Context Modeling module (SCM) and the High-Frequency Information Guidance (HFIG). 
To enable a fair comparison, the choice of hyperparameters and training strategy for all subsequent models remains entirely consistent with those of the Baseline.

\textbf{Effectiveness of Components.} As illustrated in Table \ref{tab:ablation}, by integrating information from multiple scanning directions at each spatial position, SCM constructs richer spatial context representations than the backbone alone, resulting in consistent improvements across all evaluation metrics.
Introducing HFIG also brings improvements, indicating that frequency-domain structural cues provide more robust visual representations under appearance variations. Although the improvement is relatively small, HFIG effectively complements SCM by enhancing stable structural information.
Combining SCM and HFIG yields the optimal performance simultaneously, as observed in our SLGeo. These experimental results successfully validate the effectiveness of our proposed HFIG and SCM.

\begin{table}[t]
\centering
\resizebox{\linewidth}{!}{
\setlength{\tabcolsep}{3pt} 
\begin{tabular}{lcccccccc}
\toprule
\multirow{2}{*}{Methods} & \multicolumn{4}{c}{CVACT\_val} & \multicolumn{4}{c}{CVACT\_test} \\
\cmidrule(lr){2-5} \cmidrule(lr){6-9}
 & R@1 & R@5 & R@10 & R@1\% & R@1 & R@5 & R@10 & R@1\% \\
\midrule
Baseline & 85.37 & 94.83 & 96.23 & 98.50 & 66.50 & 87.94 & 90.08 & 98.49 \\
Baseline + SCM & 90.21 & 96.07 & 96.99 & 98.47 & 71.90 & 91.71 & 93.73 & 98.36 \\
Baseline + HFIG & 89.80 & 95.69 & 96.67 & 98.30 & 71.72 & 91.68 & 93.46 & 98.32 \\
Baseline + SCM + HFIG (Ours) & \textbf{91.02} & \textbf{96.83} & \textbf{97.53} & \textbf{98.86} & \textbf{72.28} & \textbf{92.69} & \textbf{94.47} & \textbf{98.69} \\
\bottomrule
\end{tabular}
}
\caption{Ablation study on CVACT dataset. We evaluate the Effectiveness of each component in our method. Best results are in bold.}
\label{tab:ablation}
\end{table}

\begin{table}[t]
\centering
\resizebox{\linewidth}{!}{
\small
\setlength{\tabcolsep}{3pt}
\begin{tabular}{lcccccccc}
\toprule
\multirow{2}{*}{strategie} & \multicolumn{4}{c}{CVACT\_val} & \multicolumn{4}{c}{CVACT\_test} \\
\cmidrule(lr){2-5} \cmidrule(lr){6-9}
 & R@1 & R@5 & R@10 & R@1\% & R@1 & R@5 & R@10 & R@1\% \\
\midrule
Repeated Scan & 90.14 & 96.38 & 97.29 & 98.66 & 71.01 & 91.68 & 93.68 & 98.40 \\
Bidirectional Scan & 90.29 & 95.94 & 96.92 & 98.55 & 71.34 & 91.86 & 93.95 & 98.45 \\
Cross Scan & 90.38 & 96.06 & 97.01 & 98.45 & 71.84 & 91.69 & 93.66 & 98.36 \\
Rotation Scan (Ours) & \textbf{91.02} & \textbf{96.83} & \textbf{97.53} & \textbf{98.86} & \textbf{72.28} & \textbf{92.69} & \textbf{94.47} & \textbf{98.69} \\
\bottomrule
\end{tabular}
}
\caption{Effectiveness of scanning strategies in SCM. Best results are in bold.}
\label{tab:ablation2}
\end{table}

\begin{figure}[t]
  \centering
  \includegraphics[width=\linewidth]{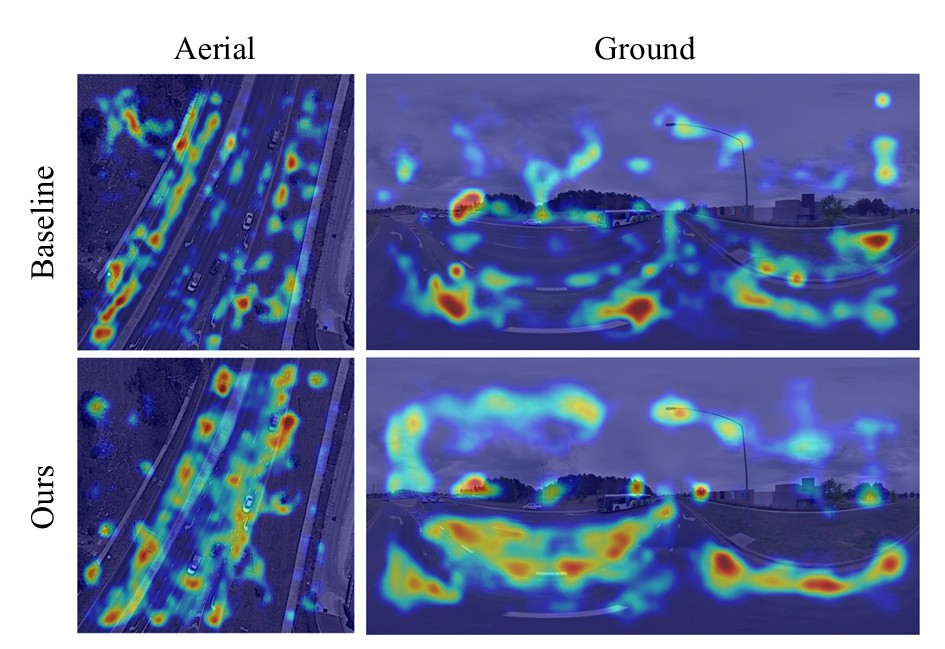}
  \caption{Heatmap visualization of the Baseline and SLGeo model. Best viewed on screen with zoom-in.}
  \label{heatmap}
\end{figure}

\textbf{Effectiveness of Scanning Strategies.} We further investigate the impact of different scanning strategies in SCM, and the results are summarized in Table \ref{tab:ablation2}. 
Specifically, Repeated Scan performs repetitive row-major scanning four times, Bidirectional Scan introduces two forward and two reverse scanning directions, Cross Scan performs horizontal and vertical scanning together, and the corresponding opposite directions, while our proposed Rotation Scan constructs four scanning sequences by rotating the feature map by $0^\circ$, $90^\circ$, $180^\circ$, and $270^\circ$.

As shown in Table~\ref{tab:ablation2}, Repeated Scan achieves the lowest performance since the four identical traversal orders provide redundant observations and contribute little additional contextual information. 
Bidirectional Scan introduces reverse traversal, leading to moderate improvements by enriching the sequential context. 
Cross Scan further exploits horizontal and vertical traversals, producing more diverse observations and yielding better performance.

Rotation Scan achieves the best performance. Unlike conventional scanning strategies that only change traversal directions, Rotation Scan changes the serialization order by rotating the feature map while preserving its underlying spatial structure.
Consequently, each spatial position is represented under four complementary directional observations, enabling SCM to aggregate richer spatial context through inter-direction interaction. These complementary observations provide a more complete representation of the scene layout, leading to improved cross-view matching performance.

\textbf{Visualization Analysis.} As shown in Figure~\ref{heatmap}, we visualize heatmaps to comprehend what SLGeo has learned and compare the differences in the regions that different models focus on.
The baseline mainly focuses on discriminative local regions, while SLGeo exhibits a noticeably broader attention distribution covering roads, intersections, surrounding vegetation, and other structural scene elements. 
Rather than relying primarily on isolated objects, SLGeo encourages integrating complementary contextual information distributed across the entire scene.
This observation is consistent with the design of SLGeo. By aggregating representations from multiple scanning directions, SCM enriches the spatial context available at each location, resulting in a more comprehensive scene understanding. 
This also explains the robustness of SLGeo under limited FoV settings. By exploiting complementary contextual information beyond a few salient regions, the learned representation remains effective even when part of the scene is missing.

\section{Conclusion}
In this work, we propose SLGeo, a novel cross-view geo-localization method that enhances scene understanding through holistic spatial layout representation. Specifically, we introduce a Spatial Context Modeling (SCM) module, which aggregates complementary observations to construct richer spatial context representations via position-wise multi-directional interaction. 
Furthermore, a High-Frequency Information Guidance (HFIG) is designed to exploit frequency-domain structural cues to anchor stable visual representation. 
Extensive experiments on CVUSA and CVACT demonstrate that SLGeo achieves competitive performance. Notably, SLGeo exhibits strong robustness under challenging conditions with unknown orientation and limited FoVs.

\bibliography{aaai2027}


\end{document}